\pdfoutput=1

\documentclass[11pt]{article}

\usepackage{EMNLP2023}

\usepackage{times}
\usepackage{latexsym}
\usepackage[capitalize,noabbrev]{cleveref}

\usepackage{relsize,etoolbox}

\AtBeginEnvironment{quote}{\fontsize{10}{11}\selectfont}

\usepackage[T1]{fontenc}

\usepackage[utf8]{inputenc}

\usepackage{microtype}
\usepackage{booktabs}
\usepackage{inconsolata}
\usepackage{lipsum} 

\usepackage{xcolor, colortbl}
\usepackage{multirow}
\usepackage{graphicx}  
\usepackage{float}
\usepackage{tabularx}
\usepackage{multirow}
\usepackage{enumitem}

  {\list{}{\leftmargin=0.1in\rightmargin=0.1in}\item[]}%
  {\endlist}

\definecolor{greenish}{HTML}{83E4B5}

\title{Answerability in Retrieval-Augmented \\ Open-Domain Question Answering}

\author{
  Rustam Abdumalikov \\
  University of Tartu \\
  rustam.abdumalikov.cs@gmail.com \\\And
  Pasquale Minervini \\
  University of Edinburgh \\
  p.minervini@ed.ac.uk \\\And
  Yova Kementchedjhieva \\
  University of Copenhagen \\
  yova@di.ku.dk
}

\begin{document}
\maketitle
\begin{abstract} 

The performance of Open-Domain Question Answering (ODQA) retrieval systems can exhibit sub-optimal behavior, providing text excerpts with varying degrees of irrelevance. Unfortunately, many existing ODQA datasets lack examples specifically targeting the identification of irrelevant text excerpts. Previous attempts to address this gap have relied on a simplistic approach of pairing questions with random text excerpts. This paper aims to investigate the effectiveness of models trained using this randomized strategy, uncovering an important limitation in their ability to generalize to irrelevant text excerpts with high semantic overlap. As a result, we observed a substantial decrease in predictive accuracy, from $98\%$ to $1\%$. To address this limitation, we discovered an efficient approach for training models to recognize such excerpts. By leveraging unanswerable pairs from the SQuAD 2.0 dataset, our models achieve a nearly perfect ($\approx 100\%$) accuracy when confronted with these challenging text excerpts.




{\color{blue}}
{\color{red}}

\end{abstract}

\section{Introduction}
Open-domain Question Answering (ODQA) systems are a powerful and widely sought-after Natural Language Processing tool. Recognizing the limits to the memory capacity of a single model and the ever-evolving nature of knowledge, many such systems rely on retrieval augmentation, wherein answers are sourced from text excerpts retrieved from reliable sources such as Wikipedia, news outlets, etc. In a prolific body of research on the topic of ODQA, one aspect that is consistently overlooked is the handling of cases where the retrieved context does not provide sufficient information to form an answer to the given question. Models trained to \emph{always} provide an answer lack the ability to abstain from doing so, 
instead of being forced to guess the answer, 
which confounds true model errors with deficiencies in the retrieval process.

\citet{kwiatkowski-etal-2019-natural} discuss this issue when designing the now widely used ODQA benchmark Natural Questions (NQ) and include \emph{unanswerable} questions. The general format of a training instance in NQ is a question paired with a gold-standard text excerpt relevant to the question and an exact answer pulled from the text excerpt. The authors source questions from real-world Google Search queries and employ annotators to collect relevant text excerpts and answer annotations. In the data collection process, they discovered that no relevant text excerpts could be found for 50.5\% of the initial seed of questions, and no straightforward answer could be formed based on available text excerpts for a further 14.7\% of the questions -- note that the text excerpts are said to contain an answer, but of a kind that cannot be simply stated following the guidelines. The authors choose to include these questions in the benchmark, labeling them as unanswerable, to promote the study of unanswerability in ODQA. 

There are two shortcomings in this data setup that have understandably prevented the further study of answerability on NQ (and on other ODQA datasets, which do not include unanswerable questions at all). Firstly, the \emph{unanswerable} label in NQ conflates two types of challenges in QA: (I)~the lack of relevant text excerpts for a given question, and (II)~the special case of questions for which an answer cannot be easily formulated even with reference to very relevant documents. 
Here, we focus on type I unanswerability, which arises when retrieval systems are inaccurate or when the knowledge available in the corpus is not sufficient, 
as can often be the case~\cite{karpukhin-etal-2020-dense,DBLP:journals/tmlr/IzacardCHRBJG22}. 

Second, and more important, is the matter of training models to recognize when the retrieved text is irrelevant. By design, retrieval systems rank the documents from a database with reference to a query and retrieve the 
$k$ most relevant ones. This means that ODQA models \emph{always} receive one or more text excerpts as part of the input; hence their ability to abstain from answering has to be trained and evaluated with reference to some negative samples. The type I data points in NQ do not contain such negative samples, which renders them unusable for the practical study of unanswerability. 

\citet{disentangling} propose a simple approach to address this issue---pairing questions with text excerpts sampled randomly from the complete pool of excerpts associated with NQ and labeling those questions as unanswerable. When training and evaluating data points designed in this way, they observe astonishingly high rates of models correctly abstaining from an answer (in the range of 99\%). Here, we raise the concern that this approach likely results in a simple heuristic being learned by models trained in this way, wherein abstaining from an answer amounts to recognizing that the text excerpt bears no relevance to the question, for example, based on the lack of semantic overlap. Meanwhile, in real-world scenarios, retrieved excerpts are likely to share semantic content with the question, 
while still not necessarily containing an answer to the question. 

In this work, we study the predictions of models faced with questions paired with semantically related but practically irrelevant text excerpts with the following research questions in mind:
%
\begin{itemize}[leftmargin=*]
    \item What training strategy can be used to achieve high efficiency in recognition of semantically related but practically irrelevant text excerpts and abstaining from providing answers?

    \item Are models more likely to hallucinate an answer or misguidedly extract one, falling toward confirmation bias, wherein mentions of the factually correct answer in a practically irrelevant context get extracted more often than entities of the same type appearing in such a context?
\end{itemize}
To automatically generate text excerpts according to a given set of criteria, we use ChatGPT~\footnote{\url{https://openai.com/blog/chatgpt}} to generate text excerpts accordingly for each research question above.

\section{Related Work}

In a machine reading comprehension (MRC) setup, \citet{rajpurkar-etal-2018-know} crowd-source high-quality unanswerable questions, presenting annotators with a context and having them write unanswerable questions based on it. 
%
%
%
\citet{asai-choi-2021-challenges} analyze unanswerability as part of the dataset creation process, where annotators mark a question as unanswerable if no answer is available among the retrieved documents presented to annotators.
%
%
\citet{disentangling} aim to disentangle parametric and contextual knowledge in ODQA models and try to address the cases where the model should abstain from answering when no relevant answers are present in the input contexts.
To this end, they pair questions with excerpts from the complete pool of excerpts and label such questions as unanswerable.
\citet{rajpurkar-etal-2018-know} added unanswerable questions to the SQuAD dataset~\citep{rajpurkar-etal-2016-squad}, providing a useful resource for identifying unanswerable questions.
\citet{yatskar-2019-qualitative} found that the unanswerable questions in SQuAD 2.0 are often cases of entity salads, false premises, topic errors, and missing information and are generally easy to detect.
%
%
%
%

\section{Methodology}
%
%
\paragraph{Datasets}
We rely on the version of NQ created by \citet{disentangling}, as it possesses several features of use to us. \citet{disentangling} take the answerable portion of NQ (questions with relevant excerpts and answers) and augment it in two ways. First, they add answerability data points, where questions are paired with random contexts, and the target answer is set to \emph{unanswerable}. Second, they create a counterfactual version of the questions, wherein the answer is replaced with another entity of the same type, and so is its mention in the relevant text excerpt. As the latter procedure cannot be applied to all questions, the resulting training split contains less counterfactual than factual data points, while in the evaluation set, they keep only those factual data points for which a counterfactual counterpart exists, such that the two subsets are matched in size (1,365 data points in each). 





%
\paragraph{Models}
We use T5-Large and T5-XL~\citep{DBLP:journals/jmlr/RaffelSRLNMZLL20} in all experiments reported below. This choice falls in line with the work of \citet{disentangling}, who experiment with T5-Large and T5-XXL and find that the two models exhibit similar patterns in evaluations of answerability and in terms of gains from counterfactual training. Likewise, in our case, both models show similar patterns.
Thus we reported T5-XL results in the main paper and T5-Large in \cref{sec:t5-large-results}.
%
%
We fine-tune the model on the union of factual and answerability training data from \citet{disentangling}~\footnote{We do not use counterfactual data for training as we did not find it to make a difference to answerability in any of the experimental settings.} -- hyperparameters are available in \cref{app:hyper}.

%
\paragraph{Generating Excerpts with ChatGPT}
We use ChatGPT to generate text as needed for the study of each of our research questions. We choose to do so as a cheap and scalable alternative to employing human annotators to filter through retrieved text excerpts or write them from scratch. 

All prompts given to the model share the following part ``Write a short paragraph of one to two sentences that looks factual but could be made up.'' The text generated with this prompt falls within the 75-percentile of excerpt lengths found in NQ. The different continuations to the prompt are detailed as relevant in subsequent sections.

In the course of developing the different prompts, we manually inspected model outputs to ensure their good quality. After settling on the final form of a prompt, we quantified how well ChatGPT followed our instructions as the rate at which entities we instructed the model to include in the generated text were indeed included. Throughout subsequent sections, this success rate is reported as $SR=96\%$.

Synthetic excerpts are generated for evaluation purposes only, i.e. they are based on questions from the evaluation split of the data paired with their factual answers unless otherwise stated. In \cref{sec:prompting}, we provide examples of the text generated with each of the prompts discussed below.

\begin{table}
\begin{tabularx}{\columnwidth}{X}
\toprule
\fontsize{10}{11}\selectfont\textbf{Question:} Who is the actor who plays King Joffrey? \\
\midrule
\fontsize{10}{11}\selectfont\textbf{Related to Q:} Certainly, King Joffrey's character in Game of Thrones was very memorable and was brought to life convincingly by an actor whose work was highly appreciated by the fans. It's interesting how a performance can sometimes transcend the character and leave a lasting impression on the audience. \\
\midrule
\fontsize{10}{11}\selectfont\textbf{Related to A:} Jack Gleeson walked through the bustling streets of Dublin, his hood pulled up to shield himself from the paparazzi, who constantly hounded him since his sudden retirement from acting. Once he reached his apartment, he settled down to paint, letting the colors and brushstrokes tell the stories he could no longer act out. \\

\bottomrule
\end{tabularx}
\caption{Contexts generated with ChatGPT.}
\label{tab:chatgpt_examples}
\end{table}

\section{RQ1: Abstaining from Answering}

\citet{disentangling} propose to train models to abstain from answering with a data augmentation wherein questions are paired with random text excerpts, and models have to predict the word ``unanswerable'' instead of an actual answer.

We aim to determine how well a model trained in this way would generalize to more challenging and realistic cases, where the excerpt is not random but rather related to the question without giving an answer to it.  To this end, we generate text excerpts with ChatGPT using two prompts; the first is meant to challenge the model by mentioning relevant entities from the question in a different context (row \textit{Related to Q} in \cref{tab:chatgpt_examples}):

\begin{quote}
    ... You must NOT give a direct answer to the question $\langle$question$\rangle$. But you must discuss the question while not mentioning the word(s) $\langle$answer$\rangle$.
\end{quote}
\noindent and the second meant to challenge the model by mentioning the answer in a different context (row \textit{Related to A} in \cref{tab:chatgpt_examples}):

\begin{quote}
    ... You must mention the word(s) $\langle$answer$\rangle$. But you must NOT discuss the question $\langle$question$\rangle$ or give an answer to it.
\end{quote}

In \cref{tab:RQ1}, we present the rate at which our T5 model, after fine-tuning, accurately refrains from answering unanswerable questions. This evaluation is conducted using data from \citet{disentangling} on one hand and our ChatGPT-augmented data points on the other. 

Similarly to \citet{disentangling}, we observe almost ceiling performance on data points with random excerpts. Meanwhile, when dealing with semantically related excerpts, the model fine-tuned on factual and answerability data points exhibits a significant decrease in its ability to abstain from answering. The rate of abstention drops to just 1.1\%. Interestingly, even with excerpts that are not directly related to the question but mention the correct answer in some form, the rate of abstaining from an answer decreases by 30\% compared to the random-excerpt setting. This hints at a possible confirmation bias\footnote{Confirmation bias is a tendency, originally observed in humans, to favor information that aligns with, i.e. confirms, prior beliefs.} in the model, wherein its prior knowledge of the correct answer appears to be interfering with its ability to abstain from answering.


\begin{table}[]
    \centering
    \resizebox{\columnwidth}{!}{
    \begin{tabular}{lccc}
    \toprule
        {\bf Data} & {\bf Random} & {\bf Related to Q} & {\bf Related to A} \\
        \midrule        
        \multicolumn{4}{c}{T5-XL}\\
        \midrule
        Factual+Answerability & 98.0 & 1.1 & 68.1 \\
        Factual+Answerability+\textbf{SQuAD} & 98.3 & 99.9 & 100.0 \\
    \bottomrule
    \end{tabular}
    }
    \caption{Percentage rate of abstaining from an answer on unanswerable data points with random excerpts and with semantically related excerpts. T5-Large results can be found in \cref{sec:t5-large-results} Table \ref{tab:RQ1-t5-large}.}
    \label{tab:RQ1}
\end{table}

Despite the confirmation bias possibility, it is apparent that a randomizing strategy is an inefficient approach. This could be due to the fact that such a strategy leads to low relevance between questions and text excerpts, and correspondingly a simple heuristic is needed to identify them. To address this, we opted for the opposite extreme of training on data points with high semantic overlap by utilizing unanswerable questions from the SQuAD 2.0 dataset.



In the \textit{SQuAD} row of Table \ref{tab:RQ1}, we show that the inclusion of unanswerable questions from SQuAD 2.0 results in flawless performance across various types of irrelevant text excerpts. Despite the claim of \citet{kwiatkowski-etal-2019-natural} that detecting SQuAD 2.0 unanswerable questions should be trivial, the ability of our model to do so generalizes effectively to the highly relevant text excerpts we designed.

\section{RQ2: Hallucinate or Extract}
%
%
In many cases, ODQA models are trained on datasets that do not consider the case that a question cannot be answered given the available evidence -- in the following, we investigate what kind of answers such models produce when they are not provided with the necessary evidence to answer a question.
To this end, we generate a third set of excerpts, which combines the properties of the two previous sets by explicitly discussing both the question and the entity corresponding to the answer, but without any claim of the two being related. We obtain these excerpts with the following prompt ($SR=95.9\%$):\footnote{See \cref{sec:prompting} for more details about our prompts.}

\begin{quote}
    ... You must NOT give a direct answer to the question $\langle$question$\rangle$. But you must discuss the question while not mentioning the word(s) $\langle$answer$\rangle$. And smoothly transition to discuss the word(s) $\langle$answer$\rangle$ as a continuation of the same paragraph.
\end{quote}
In the analysis of model answers in this adversarial setting, we first focus on whether they were extracted from the available text excerpts or were entirely hallucinated. We establish that by looking for an exact match of the answer in the excerpt. The results, reported in \cref{tab:RQ2} (row \textit{Factual}), indicate that the rate of extraction is high (70.7 + 2.4 = 73.1), which is in line with the way the model has been trained to form an answer. Still, the rate of hallucination is non-trivial (26.5). We manually inspect a subset of these hallucinated answers and find no clear patterns among them, instead concluding that they are the result of the model having no good strategy to handle the complex scenarios we present it with.


We further break down extracted answers into cases where the model extracted the correct factual answer or something else. We see that the correct factual answer is extracted more than 28 times as often as other entities in the context, which again points to a possible confirmation bias. We investigate this matter further by replacing the mention of a factual answer in the excerpts with a counterfactual one, sourced from \citet{disentangling}. 
The results, reported in Table~\ref{tab:RQ2}, row \textit{C-factual}, show that when the same position in the excerpt is occupied by a counterfactual answer instead of a factual one, the model extracts the entity in that position 10 percent less often. This serves to show that the model indeed exhibits a confirmation bias, preferring to extract familiar answers.

\begin{table}[]
    \centering
    \resizebox{.95\columnwidth}{!}{
    \begin{tabular}{lcccc}
    \toprule
        &\multicolumn{2}{c}{\textbf{Extracted}} & \textbf{Hallu}. & \textbf{Abstain} \\
        \cmidrule(lr){2-3}
        &\textbf{Correct} & \textbf{Other} \\
        \midrule
        &\multicolumn{3}{c}{T5-XL}\\
        \midrule
        Factual & 70.7 & 2.4 & 26.5 & 0.4 \\
        C-factual & 60.2 & 3.2 & 35.8 & 0.8 \\
    \bottomrule
    \end{tabular}
    }
    \caption{Rate at which the model extracts an answer from the excerpt, hallucinates, or abstains from answering in a setting of hard unanswerable data points. T5-Large results can be found in \cref{sec:t5-large-results} Table \ref{tab:RQ2-t5-large}}
    \label{tab:RQ2}
\end{table}

\section{Conclusions}


Retrieval systems can behave sup-optimally and provide relevant documents without an answer to a given question -- it is important for models to recognize when this happens and abstain from answering.
However, not all ODQA datasets contain unanswerable questions. 
To address this challenge, we explored the strategy of text excerpt randomization, finding that this approach can fail to generalize effectively. 
To overcome this limitation, we incorporated unanswerable questions from the SQuAD 2.0 dataset. 
By training our models on such data points, we were able to achieve near-perfect performance on ChatGPT-generated texts of varying relevance. This work aims to emphasize the importance of recognizing and abstaining from answering as a critical problem in developing a trustworthy question-answering system. 

\bibliography{emnlp2023,custom}
\bibliographystyle{acl_natbib}

\clearpage

\appendix

\section{T5-Large Results}
\label{sec:t5-large-results}

\begin{table}[!htb]
    \centering
    \resizebox{\columnwidth}{!}{
    \begin{tabular}{lccc}
    \toprule
        {\bf Data} & {\bf Random} & {\bf Related to Q} & {\bf Related to A} \\
        \midrule
        \multicolumn{4}{c}{T5-Large}\\
        \midrule
        Factual+Answerability & 98.3 & 0.8 & 78.2  \\
        Factual+Answerability+\textbf{SQuAD} & 98.5 & 99.9 & 100.0\\
    \bottomrule
    \end{tabular}
    }
    \caption{Percentage rate of abstaining from an answer on unanswerable data points with random excerpts and with semantically related excerpts.}
    \label{tab:RQ1-t5-large}
\end{table}

\begin{table}[!htb]
    \centering
    \resizebox{.95\columnwidth}{!}{
    \begin{tabular}{lcccc}
    \toprule
        &\multicolumn{2}{c}{\textbf{Extracted}} & \textbf{Hallu}. & \textbf{Abstain} \\
        \cmidrule(lr){2-3}
        &\textbf{Correct} & \textbf{Other} \\
        \midrule
        &\multicolumn{3}{c}{T5-Large}\\
        \midrule
        Factual & 68.1 & 2.4 & 29.3 & 0.2\\ 
        C-factual & 58.0 & 4.1 & 37.7 & 0.2\\
    \bottomrule
    \end{tabular}
    }
    \caption{Rate at which the model extracts an answer from the excerpt, hallucinates, or abstains from answering in a setting of hard unanswerable data points.}
    \label{tab:RQ2-t5-large}
\end{table}

\section{Hyperparameters} \label{app:hyper}

 For fine-tuning, we adopt the hyperparameters used in \citet{disentangling}: batch size 32, learning rate 0.0001, maximum sequence length 256 for training, and 396 for evaluation.

\section{Prompting ChatGPT}
\label{sec:prompting}

We suffix any prompts that explicitly ask ChatGPT to discuss a given answer with the following sequence:

\begin{quote}
 Always use [$\langle$answer$\rangle$] without quotes or brackets. If $\langle$answer$\rangle$ is a number then be creative when you discuss it. Never use words ``number'' and ``digit''.
\end{quote}
This is necessary because we used square brackets inside the prompt to emphasize the answer, and ChatGPT correspondingly tended to enclose the answer in quotes or brackets. To circumvent this, we explicitly instructed ChatGPT to refrain from such behavior.

The objective of the latter section of the prompt was to enhance the variety in the generated output. Without it, we noticed that ChatGPT consistently began the generation with the repetitive pattern of "This number/digit represents...".

\end{document}